\documentclass{article}
\usepackage{spconf,amsmath,graphicx,listings,enumitem,amsmath,booktabs,multicol,placeins,url,hyperref,float,afterpage}
\usepackage{color}
\usepackage[left=2cm, right=2cm, top=2cm, bottom=2cm]{geometry}
\usepackage{comment}
\usepackage{url}
\usepackage{fancyhdr}

\title{Camera Calibration Through Geometric Constraints from Rotation and Projection Matrices}
\name{Muhammad Waleed, Abdul Rauf, Dr. Murtaza Taj}
\address{
  Computer Vision and Graphics Lab, LUMS \\
}

\fancypagestyle{firstpage}{
  \fancyhf{} 
  \fancyfoot[L]{This work has been submitted to the IEEE for possible publication. Copyright may be transferred without notice, after which this version may no longer be accessible.} 
}

\begin{document}
%
\maketitle
\thispagestyle{firstpage}
\begin{abstract}
\noindent The process of camera calibration involves estimating the intrinsic and extrinsic parameters, which are essential for accurately performing tasks such as 3D reconstruction, object tracking and augmented reality. In this work, we propose a novel constraints-based loss for measuring the intrinsic (focal length: $(f_x, f_y)$ and principal point: $(p_x, p_y)$) and extrinsic (baseline: ($b$), disparity: ($d$), translation: $(t_x, t_y, t_z)$, and rotation specifically pitch: $(\theta_p)$) camera parameters. Our novel constraints are based on geometric properties inherent in the camera model, including the anatomy of the projection matrix (vanishing points, image of world origin, axis planes) and the orthonormality of the rotation matrix. Thus we proposed a novel Unsupervised Geometric Constraint Loss (UGCL) via a multitask learning framework. Our methodology is a hybrid approach that employs the learning power of a neural network to estimate the desired parameters along with the underlying mathematical properties inherent in the camera projection matrix. This distinctive approach not only enhances the interpretability of the model but also facilitates a more informed learning process. Additionally, we introduce a new CVGL Camera Calibration dataset, featuring over \( 900 \) configurations of camera parameters, incorporating \(63,600 \) image pairs that closely mirror real-world conditions. By training and testing on both synthetic and real-world datasets, our proposed approach demonstrates improvements across all parameters when compared to the state-of-the-art (SOTA) benchmarks. The code and the updated dataset can be found here: \url{https://github.com/CVLABLUMS/CVGL-Camera-Calibration}.

\end{abstract}
\section{Introduction}
\label{sec:intro}

\noindent Camera calibration is a process used in computer vision to determine the parameters of a camera model. The main objective of calibration is to estimate both intrinsic and extrinsic parameters of the camera. Intrinsic parameters are unique to the structure of the camera. Consist of five values including focal length (\(f_x, f_y\)), optical center (\(p_x, p_y\)) and lens distortion. 
Extrinsic parameters describe how the camera is positioned and oriented about a reference coordinate system in the world comprising six values that encompass translation $(t_x, t_y, t_z)$ and rotation $(\theta_r, \theta_p, \theta_y)$. In recent literature \cite{BARRETO2006208, NITSCHKE2011835, wilczkowiak:inria-00524415, DBLP:journals/corr/WorkmanZJ16, Zhang2020DeepPTZDS} many end-to-end learning frameworks has been proposed that directly estimate these desired parameters while disregarding underlying mathematical foundations. The only exception is Camera Calibration via Camera Projection Loss (CPL) which incorporates 3D reconstruction loss~\cite{9746819} in addition to mean absolute error on regressed parameters.  \\

\begin{figure}[t]
  \centering
  \includegraphics[width=0.5\textwidth]{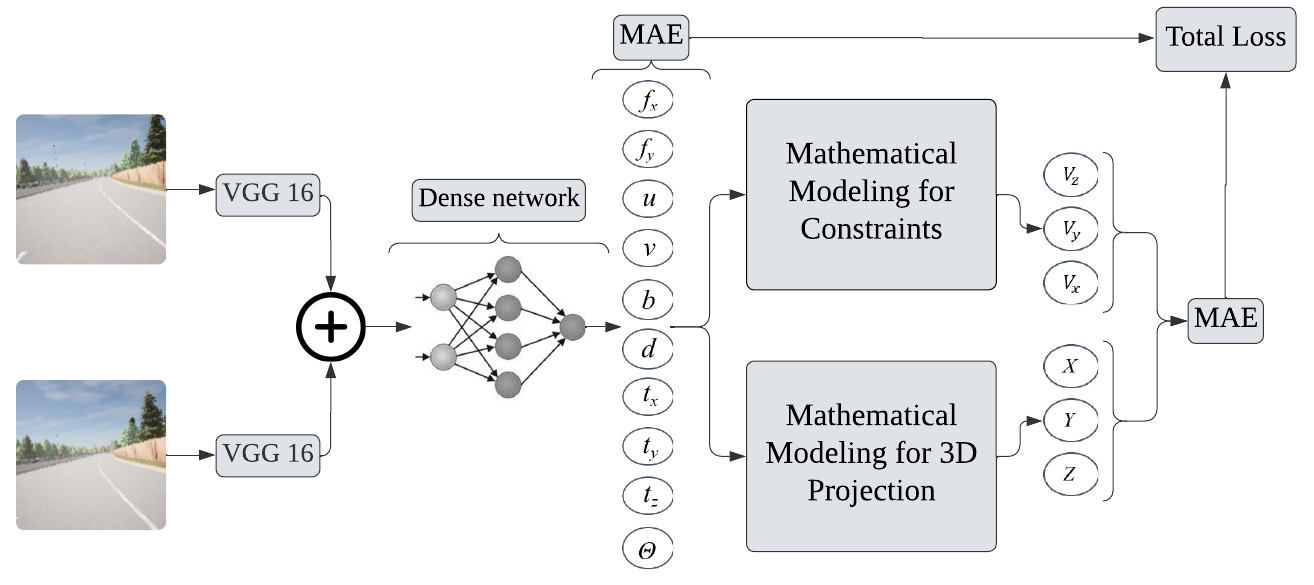}
  \caption{Constrained Approach: Leveraging the underlying mathematical operations inherent in camera model.}
  \label{fig:Architec}
\end{figure}

Our paper introduces a novel dimension to camera projection methods by incorporating general properties of the camera projection matrix as constraints. The key contributions of our work are as follows:

\begin{itemize}[label={\textbullet}]

  \item We propose a novel Unsupervised Geometric Constraint Loss (UGCL) which incorporates \( 12 \) additional constraints. These include 7 constraints from the projection matrix (3 for vanishing points, 1 for the image of world origin and 3 for axis planes orthogonality), and 5 orthonormality constraints from the rotation matrix.

  \item While CPL~\cite{9746819} is a purely supervised method, ours is a semi-supervised approach as proposed constraints do not require any additional data annotations.

  \item Our constraints enforce the inherent mathematical properties of the camera model. This aspect highlights the model's adeptness, enabling it to deliver accurate and reliable outcomes by aligning closely with the theoretical foundations of the multi-view geometry.

  \item By evaluation on unseen Daimler Stereo dataset \cite{7535515}, we show that despite training only on synthetic dataset, our method improves generalization on unseen real datasets.

  \item Beyond performance gains, our methodology contributes to the interpretability of the model. Constraint loss aids in making the learning process more transparent, allowing researchers and practitioners to gain insights into how the model makes decisions and fostering a deeper understanding of its inner workings.

  \item Furthermore, our research introduces a new CVGL Camera Calibration dataset which includes improved image quality via photorealistic rendering via CARLA Simulator. Furthermore, it has more than 900 configurations of camera parameters, incorporating approximations that closely mirror real-world conditions. Our dataset is publicly available at: \href{https://github.com/CVLABLUMS/CVGL-Camera-Calibration}{CVGL-Dataset link}.
\end{itemize}

\section{Related Work}
\label{sec:format}
Camera calibration plays a role in computer vision and photogrammetry allowing us to interpret and understand images captured by cameras accurately. It involves determining the workings and positioning of a camera, which helps convert 3D world coordinates into 2D image coordinates. Over the years researchers have proposed methodologies and techniques to overcome the challenges tied to camera calibration.\\

It's worth noting that most existing studies \cite{BARRETO2006208, DBLP:journals/corr/DeToneMR16, NITSCHKE2011835, wilczkowiak:inria-00524415, DBLP:journals/corr/WorkmanZJ16, Workman2015DEEPFOCALAM, Zhang2020DeepPTZDS} have primarily focused on subsets of camera parameters overlooking the estimation of the set through multi-task learning. Such as, ``DeepFocal'' \cite{Workman2015DEEPFOCALAM} presents an adaptation of the AlexNet \cite{krizhevsky2012imagenet} architecture, for estimating the horizontal field of view from images. By transforming the architecture into a regression model with a single output node, the method demonstrates a specialized application of deep learning for precise field-of-view estimation. ``DeepHomo'' \cite{DBLP:journals/corr/DeToneMR16}, takes two images and estimates the relative homography using a deep convolutional neural network. It has a network of \(10\) layers that takes a stack of greyscale images and outputs the homography matrix (\(8\) degrees of freedom) that can be used to map the pixels from one image to another.\\

Camera Projection Loss (CPL) \cite{9746819} represents an advancement in this field. Their paper simultaneously estimates both intrinsic and extrinsic camera parameters using a multi-task learning framework. CPL utilized the projection of 2D image coordinates \((x, y)\) onto world coordinates \((X, Y, Z)\) as a reference point which serves as a proxy measure, for estimating both intrinsic and extrinsic parameters. ``NeRFtrinsic Four'' \cite{Schieber2023NeRFtrinsicFA}, instead perform camera calibration as a subtask for novel view synthesis from different viewpoints. In previous methods for novel view synthesis, estimation of intrinsic and extrinsic parameters was required. So, instead of relying on the prior information of these parameters, they estimate calibration parameters using Gaussian Fourier Feature Mapping. In ``Template Detection'' paper \cite{Dantas2022AutomaticTD}, proposes a two-step convolutional neural network framework for automatic checkerboard detection and corner point detection, by identifying the corner points, this method provides essential points for the camera calibration. Unlike our proposed approach, none of these methods rely on constraints inherently present in camera projection and rotation matrices.

\section{Our Methodology}
\label{sec:pagestyle}

In our proposed approach, we took inspiration from the CPL framework~\cite{9746819} to compute a complete camera projection model. We employ a multi-task learning framework that utilizes dependent regressors sharing a common feature extractor. To be specific we make use of an Inception v3 \cite{7780677} model that has been pre-trained on ImageNet \cite{deng2009imagenet} as the feature extractor and incorporates mathematical layers for calculating loss. In previous work, $13$ regressors were used, with $10$ corresponding to extrinsic camera parameters and $3$ corresponding to the 3D point cloud.\\

Building upon this foundation our extension involves incorporating constraints related to general camera anatomy as additional parameters. To ensure these crucial constraints are met we introduce constraint loss terms for each of these parameters within the total loss function. This extension aims to improve the model’s capability in utilizing geometric information for more accurate estimation of camera parameters.\\

For a pinhole camera \cite{article} a point in the 3D world coordinate system undergoes projection first into the camera coordinate system and then into the image coordinate system. This process can be represented as follows.

\begin{equation}
{\small
\begin{bmatrix}
    x' \\
    y' \\
    1
\end{bmatrix}
= 
\begin{bmatrix}
    f_x & 0 & p_x \\
    0 & f_y & p_y \\
    0 & 0 & 1
\end{bmatrix}
\begin{bmatrix}
    r_{11} & r_{12} & r_{13} & t_x \\
    r_{21} & r_{22} & r_{23} & t_y \\
    r_{31} & r_{32} & r_{33} & t_z \\
\end{bmatrix}
\begin{bmatrix}
    X \\
    Y \\
    Z \\
    1
\end{bmatrix}
}
\end{equation}

\begin{equation}
x = K [R|T] X
\end{equation}

In general, the camera model is defined as:

\begin{equation}
\resizebox{0.3\textwidth}{!}{$
\begin{bmatrix}
    x \\
    y \\
    1
\end{bmatrix}
=
\begin{bmatrix}
    a_{11} & a_{12} & a_{13} & a_{14} \\
    a_{21} & a_{22} & a_{23} & a_{24} \\
    a_{31} & a_{32} & a_{33} & a_{34}
\end{bmatrix}
\begin{bmatrix}
    X \\
    Y \\
    Z \\
    1
\end{bmatrix}
$}
\end{equation}

\begin{equation}
x = PX 
\end{equation}

In this context, $K$ represents the intrinsic matrix, $[R|T]$ represents the extrinsic matrix and $P$ represents the projection matrix. The projection matrix $P$ and rotation matrix $R$ possess properties that can be utilized as constraints in our task framework.

\subsection{Properties of Rotation Matrix:}
\label{sec:Pop Rotation}
\begin{itemize}
    \item The rotation matrix is characterized by orthogonality. This means that when we multiply any two rows of this matrix, the result is consistently zero.
    \begin{align}
    \text{$\boldsymbol{r}^{1T} \cdot \boldsymbol{r}^{2T} = 0$} \\
    \text{$\boldsymbol{r}^{1T} \cdot \boldsymbol{r}^{3T} = 0$} \\
    \text{$\boldsymbol{r}^{2T} \cdot \boldsymbol{r}^{3T} = 0$}
    \end{align}

    {where $\boldsymbol{r}^{1T}$, $\boldsymbol{r}^{2T}$, $\boldsymbol{r}^{3T}$ represent the 1st, 2nd, and 3rd rows of the rotation matrix $R$.}
    
    \item The product of the rotation matrix and its transpose always results in an identity matrix.
    \begin{equation}
    R \cdot R^\top = I_\text{identity} 
    \end{equation}

    \item The determinant of the rotation matrix is always one.
    \begin{equation}
    \det(R) = 1 
    \end{equation}
    
\end{itemize}

\subsection{Properties of Projection Matrix:}
\begin{itemize}
     \item The first column of the projection matrix $P$ symbolizes the image of the point at infinity, along the X-axis which we call the first vanishing point $V_x$. Likewise, the second column corresponds to the Y-axis representing the vanishing point $V_y$. Lastly, the third column aligns with the Z-axis. Represents the vanishing point $V_z$. This representation can be expressed using an inhomogeneous system.

     \begin{gather}
    \resizebox{0.37\textwidth}{!}{$
    \begin{aligned}
    \mathbf{V}_x &= \begin{bmatrix} a_{11}/a_{31} \\ a_{21}/a_{31} \\ a_{31}/a_{31} \end{bmatrix}, \hfil
    \mathbf{V}_y &= \begin{bmatrix} a_{12}/a_{32} \\ a_{22}/a_{32} \\ a_{32}/a_{32} \end{bmatrix},  \hfil
    \mathbf{V}_z &= \begin{bmatrix} a_{13}/a_{33} \\ a_{23}/a_{33} \\ a_{33}/a_{33} \end{bmatrix} 
    \end{aligned}
    $} \label{eq:vanishing point}
    \end{gather}

    \item The fourth column of projection $P$ represents the world center. That can be represented in an inhomogeneous coordinate system.

    \begin{equation}
    \resizebox{0.12\textwidth}{!}{$
        \mathbf{W}_c = \begin{bmatrix} a_{14}/a_{34} \\ a_{24}/a_{34} \\ a_{34}/a_{34} \end{bmatrix}
    $} \label{eq:world center}
    \end{equation}
    
    \item Each row of projection matrix $P$ represents the axis plane, leading to the property that the cross product of any two rows is zero.

    \begin{align}
    \text{$\boldsymbol{p}^{1T} \times \boldsymbol{p}^{2T} = 0$} \\
    \text{$\boldsymbol{p}^{1T} \times \boldsymbol{p}^{3T} = 0$} \\
    \text{$\boldsymbol{p}^{2T} \times \boldsymbol{p}^{3T} = 0$}
    \end{align}
    
    {where $\boldsymbol{p}^{1T}$, $\boldsymbol{p}^{2T}$, $\boldsymbol{p}^{3T}$ represent the 1st, 2nd, and 3rd rows of the projection matrix $P$.}
\end{itemize}

The constraints imposed by the rotation matrix serve as proxy variables for Pitch, Yaw and Roll. Simultaneously, the projection matrix constraints act as proxy variables for $K$, $R$, and $T$, encompassing all target parameters.

\subsection{3D reconstruction:}

A 2D point image coordinate system is first converted to a camera coordinate and then to a 3D point. So, the conversion of 2D point to camera coordinate can be written as: \\

\begin{equation}
{\small
\begin{bmatrix}
    {y}_{cam} \\
    {z}_{cam} \\
    {x}_{cam} \\
    {1}
\end{bmatrix}
= 
\begin{bmatrix}
    \frac{1}{f_x} & 0 & \frac{-p_x}{f_x} \\
    0 & \frac{1}{f_y} & \frac{-p_y}{f_y} \\
    0 & 0 & 1
\end{bmatrix}
\begin{bmatrix}
    x \\
    y \\
    1
\end{bmatrix}
}
\end{equation}

\begin{align}
  x_{\text{cam}} &= 1 \\
  y_{\text{cam}} &= \frac{x}{f_x} - \frac{p_x}{f_x} = \frac{x - p_x}{f_x} \\
  z_{\text{cam}} &= \frac{y}{f_y} - \frac{p_y}{f_y} = \frac{y - p_y}{f_y}
\end{align}

The camera to world transformation:

\begin{equation}
{\small
\begin{bmatrix}
    {X} \\
    {Y} \\
    {Z} \\
    {1}
\end{bmatrix}
= 
\begin{bmatrix}
   R & t \\
   0_{1x3} & 1
\end{bmatrix}
\begin{bmatrix}
    {x}_{cam} \\
    {y}_{cam} \\
    {z}_{cam} \\
    {1}
\end{bmatrix}
}
\end{equation}

\begin{equation}
{\small
\begin{bmatrix}
    X \\
    Y \\
    Z \\
\end{bmatrix}
= 
\begin{bmatrix}
   \cos \theta & 0 & \sin \theta \\
   0 & 1 & \theta \\
   -\sin \theta & 0 & \cos \theta
\end{bmatrix}
\begin{bmatrix}
    x_{\text{cam}} \\
    y_{\text{cam}} \\
    z_{\text{cam}} \\
\end{bmatrix}
+
\begin{bmatrix}
    t_{x} \\
    t_{y} \\
    t_{z} \\
\end{bmatrix}
}
\end{equation}

\begin{align}
  X &= x_{cam} * cos \theta + z_{cam} * sin \theta + t_x \\
  Y &= y_{cam} + t_y \\
  Z &= -x_{cam} * sin \theta + z_{cam} * cos \theta + t_z
\end{align}

Overall to project a point from a 2D image back to 3D world, we can write as:

\begin{align}
  x_{cam} &= f_x * b/d \\
  y_{cam} &= -(x_{cam}/f_x) * (x - p_x) \\
  z_{cam} &= (x_{cam}/f_y) * (p_y - y)
\end{align}

In 3D point reconstruction, $x_{cam}$ acts as a proxy variable for $f_x$, disparity and baseline, $y_{cam}$ acts as proxy variable for $f_x$, $x$ and $p_x$ while $z_{cam}$ acts as a proxy variable for $f_y$, $y$ and $p_y$. \\

\begin{table*}[ht]
  \centering
  \caption{MAE in the predicted parameters on the updated CVGL test dataset comprising 19,080 images.}
  \label{tab:your_table_label}

  \begin{tabular}{l|l|c|l|c|l|c|l|c|l|cccccc|}
    \hline
    & $f_x$ & $f_y$ & $p_x$ & $p_y$ & $b$ & $d$ & $t_x$ & $t_y$ & $t_z$ & $\theta_p$ \\
    \hline
    DeepFocal \cite{Workman2015DEEPFOCALAM} & 4.937 & 4.936 & 0.976 & 0.974 & 0.154 & 4.045 & 0.219 & 0.139 & 0.145 & 0.109 \\
    DeepHomo \cite{DBLP:journals/corr/DeToneMR16} & 6.434 & 6.434 & 0.987 & 0.988 & 0.250 & 2.855 & 0.308 & 0.196 & 0.238 & 0.206 \\
    Baseline \cite{9746819} & 9.285 & 9.157 & 1.165 & 0.816 & 14.31 & 3.091 & 20.05 & 12.51 & 12.47 & 0.088 \\
    MTL-A \cite{9746819} & 3.697 & 3.646 & 0.811 & 0.547 & 0.701 & 2.851 & 0.402 & 0.451 & 0.250 & 0.088 \\
    MTL-U \cite{9746819} & 31.13 & 33.10 & 5.492 & 5.177 & 0.663 & 10.81 & 0.274 & 0.171 & 0.179 & 0.090 \\
    UGCL-VP-WC-R (Ours) & \textbf{1.747} & \textbf{1.804} & \textbf{0.139} & \textbf{0.089} & \textbf{0.143} & \textbf{2.542} & \textbf{0.200} & \textbf{0.125} & \textbf{0.126} & \textbf{0.009} \\
    \hline
  \end{tabular}
\end{table*}

\begin{table*}[ht]
  \centering
  \caption{MAE in the predicted parameters on the Daimler Stereo \cite{7535515} test dataset comprising 5,389 images (Forward-Pass-Only).}
  \label{tab:Daimler}

  \begin{tabular}{l|l|c|l|c|l|c|l|c|l|cccccc|}
    \hline
    & $f_x$ & $f_y$ & $p_x$ & $p_y$ & $b$ & $d$ & $t_x$ & $t_y$ & $t_z$ & $\theta_p$ \\
    \hline
    DeepFocal \cite{Workman2015DEEPFOCALAM} & 57.78 & 48.36 & 8.475 & 8.950 & 0.608 & 45.03 & 1.991 & 1.552 & \textbf{0.542} & 0.737 \\
    DeepHomo \cite{DBLP:journals/corr/DeToneMR16} & 84.49 & 75.13 & 1.190 & 1.188 & 0.647 & 37.84 & 0.792 & 1.590 & 1.297 & 0.367 \\
    Baseline \cite{9746819} & 11.87 & 10.96 & 0.617 & 74.48 & 74.36 & 47.74 & 45.23 & 45.51 & 25.98 & 6.527 \\
    MTL-A \cite{9746819} & 10.94 & 9.894 & 0.455 & 0.271 & 0.125 & 4.610 & \textbf{0.253} & 2.000 & 1.670 & 0.329 \\
    MTL-U \cite{9746819} & 87.99 & 79.75 & 16.82 & 16.44 & \textbf{0.086} & 71.52 & 0.896 & 1.700 & 1.583 & 0.360 \\
    UGCL-VP-WC-R (Ours) & \textbf{10.63} & \textbf{9.691} & \textbf{0.065} & \textbf{0.070} & 0.586 & \textbf{4.362} & 0.664 & \textbf{1.534} & 1.447 & \textbf{0.257} \\
    \hline
  \end{tabular}
\end{table*}

\subsection{Unsupervised Geometric Constraint Loss}

Building upon the camera projection loss (CPL) which effectively tackles the challenge of training an architecture for parameters with varying magnitudes we enhance its loss function. Instead of optimizing camera parameters separately, we introduce an innovative approach by including constraint losses. This extension aims to refine the model's training process by not only considering the 2D to 3D projection of points as proposed in CPL but also enforcing additional constraints on crucial parameters like vanishing points, projection matrix, rotation matrix, world center, and camera center. The augmented loss function is designed to strike a balance between accurate parameter estimation and adherence to geometric principles thus improving the overall robustness and interpretability of the camera calibration model.\\

As shown in Fig.~\ref{fig:Architec}, for given stereo images, our model predicts three set of parameters: $s_1'$ = ($f_x'$, $f_y'$, $p_x'$, $p_y'$, $b'$, $d'$, $\theta_p'$, $t_x'$, $t_y'$, $t_z'$), $s_2'$ = ($X'$, $Y'$, $Z'$) and $s_3'$ = ($V_x'$, $V_y'$, $V_z'$) where each set corresponds to calibration, 3D projection and constraints, respectively. The mean absolute error (MAE) is then computed with the actual values: $s_1$ = ($f_x$, $f_y$, $p_x$, $p_y$, $b$, $d$, $\theta_p$, $t_x$, $t_y$, $t_z$), $s_2$ = ($X$, $Y$, $Z$) and $s_3$ = ($V_x$, $V_y$, $V_z$). The total loss is then computed as the sum of all the losses for each set.

\begin{equation}
    L_1(s_1', s_1) = \frac{1}{n} \sum_{i=1}^{n} MAE(s_1', s_1) \tag{17}
\end{equation}
\begin{equation}
    L_2(s_2', s_2) = \frac{1}{n} \sum_{i=1}^{n} MAE(s_2', s_2) \tag{18}
\end{equation}
\begin{equation}
    L_3(s_3', s_3) = \frac{1}{n} \sum_{i=1}^{n} MAE(s_3', s_3) \tag{19}
\end{equation}
\begin{equation}
    L_T = \frac{{L_1 + L_2 + L_3}}{3} \tag{20}
\end{equation}

Instead of directly regressing the 3D points ($X'$, $Y'$, $Z'$) and constraints ($V_x'$, $V_y'$, $V_z'$), we extended the approach by directly regressing the camera parameters and using these parameters to construct the constraints along with the 3D points. Such that, these 3D points and constraints serve as proxy variables for camera parameters.\\

To address challenges arising from predicting 3D points and constraints using camera parameters, a common issue arises when a proxy variable deviates from its ideal value it can be attributed to multiple parameters. This leads to difficulty in convergence as the loss from one parameter can impact another parameter through the total loss. In response, to this issue our approach draws inspiration from the disentangle camera projection loss technique \cite{9746819} similar to \cite{Lopez_2019_CVPR}. We expand on this method by incorporating constraint parameters mitigating error backpropagation and enhancing the model's understanding of the scene. This modified version enhances the learning process by focusing on aspects thereby improving the accuracy of camera calibration predictions.\\

\noindent Camera Parameters:
\vspace{-0.1em}

\begin{equation}
\scriptsize
\begin{aligned}
&L_{f_x} = (f_x, f_y^{GT}, u_0^{GT}, v_0^{GT}, b^{GT}, d^{GT}, \theta_p^{GT}, t_x^{GT}, t_y^{GT}, t_z^{GT}, \text{actual}) \\
&L_{f_y} = (f_x^{GT}, f_y, u_0^{GT}, v_0^{GT}, b^{GT}, d^{GT}, \theta_p^{GT}, t_x^{GT}, t_y^{GT}, t_z^{GT}, \text{actual}) \\
&\dots \\
&L_{t_z} = (f_x^{GT}, f_y^{GT}, u_0^{GT}, v_0^{GT}, b^{GT}, d^{GT}, \theta_p^{GT}, t_x^{GT}, t_y^{GT}, t_z, \text{actual})
\end{aligned} \notag
\end{equation}

\vspace{0.1em}

\begin{equation}
\small
\begin{aligned}
&L_{\text{Cam}} = \frac{{L_{f_x} + L_{f_y} + \ldots + L_{t_z}}}{10}
\end{aligned}
\end{equation}

\noindent 3D Reconstruction:

\vspace{-0.1em} 

\begin{equation}
\small
\begin{aligned}
&L_{X} = (X, Y^{GT}, Z^{GT}, \text{actual}) \\
&L_{Y} = (X^{GT}, Y, Z^{GT}, \text{actual}) \\
&L_{Z} = (X^{GT}, Y^{GT}, Z, \text{actual})
\end{aligned} \notag
\end{equation}

\vspace{0.1em}

\begin{equation}
\small
\begin{aligned}
&L_{3D} = \frac{{L_{X} + L_{Y} + L_{Z}}}{3} 
\end{aligned}
\end{equation}
\noindent Constraints:

\vspace{-0.1em} 

\begin{equation}
\small
\begin{aligned}
&L_{V_x} = (V_x, V_y^{GT}, V_z^{GT}, \text{actual}) \\
&L_{V_y} = (V_x^{GT}, V_y, V_z^{GT}, \text{actual}) \\
&L_{V_z} = (V_x^{GT}, V_y^{GT}, V_z, \text{actual}) \\
\end{aligned} \notag
\end{equation}
\vspace{0.1em}

\begin{equation}
\small
\begin{aligned}
&L_{con} = \frac{{L_{V_x} + L_{V_y} + L_{V_z}}}{3} 
\end{aligned}
\end{equation}

As it is difficult to know which constraint helps better to converge the model, we have assigned learnable parameters ``$\omega$'' with each of the output loss parameters having a sigmoid activation function to confine their values between $0$ to $1$, representing varying degrees of importance. This gives the ability to the model to select the important parameters only. The updated loss is given as:

\vspace{-0.1em}
\begin{equation}
\begin{aligned}
\small
&L_{\text{Cam}} = \frac{\omega_{1}L_{f_x} + \omega_{2}L_{f_y} + \ldots + \omega_{10}L_{t_z}}{10}\\
\vspace{0.5em}
&L_{\text{3D}} = \frac{\omega_{11}L_{X} + \omega_{12}L_{Y} + \omega_{13}L_{Z}}{3}\\
\vspace{0.5em}
&L_{\text{Con}}  = \frac{\omega_{14}L_{V_x} + \omega_{15}L_{V_y} + \omega_{16}L_{V_z}}{3}
\end{aligned}
\end{equation}

Total loss can be written as the sum of camera parameters, 3D point reconstruction, and constraints loss.

\begin{equation}
\begin{aligned}
\small
&L_{\text{total}} = \frac{\omega_{17}{L_{\text{Cam}} + \omega_{18}L_{3D} + \omega_{19}L_{con}}}{3}
\end{aligned}
\end{equation}

\section{Dataset}
We generated a new CVGL Camera Calibration dataset using Carla Simulator \cite{DBLP:journals/corr/abs-1711-03938}. The dataset has been generated using Town~3, Town~5, and Town~6 available in CARLA. The dataset contains stereo images and for each stereo image, there is their field of view (fov), pitch, yaw, and roll for rotation and $t_x$, $t_y$, and $t_z$ for translation along with disparity and baseline. This dataset comprises over 63,600 stereo RGB images, each with a resolution of 150x150 pixels. Where 10,400 images are from Town 3, 29,000 images are from Town 5 and 24,200 are from Town 6. We have generated more than 900 configurations to approximate real world conditions. \\

To underscore the robustness of our model in real-world conditions without direct training on such data, we employed the Daimler Stereo dataset \cite{7535515}, comprising 5,389 images, solely for evaluation purposes. This dataset served as an extensive test set, reflecting authentic urban environments, where the results presented are purely the outcomes of forward passes—predictions made by our model, previously trained on the CVGL dataset, thereby demonstrating its generalization capabilities.

\begin{figure}[htb]

\begin{minipage}[b]{.3\linewidth}
  \centering
  \includegraphics[width=2.3cm]{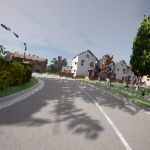}
  \centerline{(a)}
\end{minipage}
\hfill
\begin{minipage}[b]{0.3\linewidth}
  \centering
  \includegraphics[width=2.3cm]{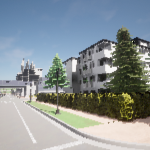}
  \centerline{(c)}
\end{minipage}
\hfill
\begin{minipage}[b]{0.3\linewidth}
  \centering
  \includegraphics[width=2.3cm]{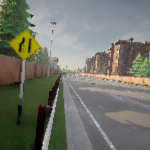} 
  \centerline{(e)}
\end{minipage}

\begin{minipage}[b]{.3\linewidth}
  \centering
  \includegraphics[width=2.3cm]{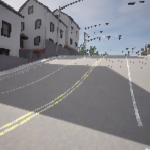}
  \centerline{(b)}
\end{minipage}
\hfill
\begin{minipage}[b]{0.3\linewidth}
  \centering
  \includegraphics[width=2.3cm]{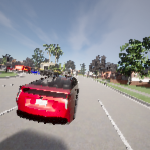}
  \centerline{(d)}
\end{minipage}
\hfill
\begin{minipage}[b]{0.3\linewidth}
  \centering
  \includegraphics[width=2.3cm]{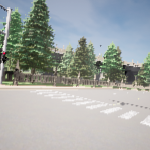} 
  \centerline{(f)}
\end{minipage}

\caption{Representative images from Town 3 (a, b), Town 5 (c, d), and Town 6 (e, f) in our synthetic dataset.}
\label{fig:res}
\end{figure}
\begin{table*}[ht]
  \centering
  \caption{Results of Ablative Study on updated CVGL test dataset comprising 19,080 images with MAE Loss}
  \label{tab:Ablative Study}

  \begin{tabular}{l|l|c|l|c|l|c|l|c|l|cccccc|}
    \hline
    & $f_x$ & $f_y$ & $p_x$ & $p_y$ & $b$ & $d$ & $t_x$ & $t_y$ & $t_z$ & $\theta_p$ \\
    \hline
    UGCL-VP & 1.979 & 1.973 & 0.334 & 0.438 & \textbf{0.143} & 2.616 & \textbf{0.200} & \textbf{0.125} & 0.126 & \textbf{0.009} \\
    UGCL-VP-WC & 1.875 & 1.900 & 0.253 & 0.129 & \textbf{0.143} & 2.640 & \textbf{0.200} & \textbf{0.125} & \textbf{0.125} & 0.013 \\
    UGCL-VP-WC-R & \textbf{1.747} & \textbf{1.804} & \textbf{0.139} & \textbf{0.089} & \textbf{0.143} & \textbf{2.542} & \textbf{0.200} & \textbf{0.125} & 0.126 & \textbf{0.009} \\
    \hline
  \end{tabular}
\end{table*}

\section{Comparison and Evaluation}
\label{sec:majhead}

For evaluation, we trained and tested every model for 100 epochs on Ge-Force RTX 2080 GPU. All the implementation is done using Keras \cite{Ketkar2017} with Mean Absolute Error (MAE) loss function and a base learning rate of 0.001 with a batch size of 32.\\

The results revealed a notable reduction in loss across parameters ($f_x$, $f_y$, $p_x$, $p_y$, $b$, $d$, $\theta_p$, $t_x$, $t_y$, $t_z$). For comparison, we used DeepHomo~\cite{DBLP:journals/corr/DeToneMR16}, DeepFocal~\cite{Workman2015DEEPFOCALAM} and CPL~\cite{9746819}. In DeepHomo and DeepFocal, we modified the output regression head to our parameters output.

\subsection{Comparison with State-of-the-Art (SOTA)}

Our approach, UGCL-VP-WC-R, stands out as a significant advancement over state-of-the-art models CPL-A and CPL-U, when trained and tested on the proposed new CVGL dataset. As showcased in Table \ref{tab:your_table_label}, our model consistently achieves lower mean absolute error (MAE) across all predicted parameters on the CVGL test dataset comprising 19,080 stereo images. Notably, for intrinsic camera parameters ($f_x$, $f_y$, $p_x$, $p_y$), our approach demonstrates substantially reduced error rates compared to both CPL-A and CPL-U, underscoring its superior capability in accurately estimating fundamental camera properties. Moreover, in terms of extrinsic camera parameters ($t_x$, $t_y$, $t_z$, $\theta_p$), UGCL-VP-WC-R consistently outperforms its counterparts, exhibiting superior accuracy in estimating translation and rotation parameters critical for stereo vision tasks. 

\begin{figure}[htb]
    \centering
    \includegraphics[width=0.8\linewidth]{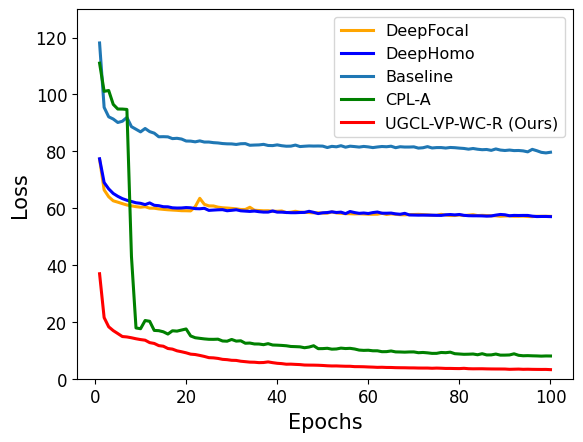}
    \caption{Comparing training losses over 100 epochs.}
    \label{fig:single}
\end{figure}

These results highlight the effectiveness of our approach in leveraging the underlying mathematics in the camera model to achieve enhanced performance. This can also be seen in Fig.~\ref{fig:single}, showing the better convergence of our approach during training over 100 epochs as compared to other state-of-the-art approaches.

\subsection{Results on Unseen data}

To assess the generalizability of our approach, we trained different architectures (shown in Table \ref{tab:Daimler}), including our own (UGCL-VP-WC-R), using the proposed CVGL dataset. We then applied these models to the unseen Daimler Stereo \cite{7535515} test dataset comprising 5,389 stereo images and computed the loss using the mean absolute error (MAE). The results, summarized in Table~\ref{tab:Daimler}, reveal promising performance. Our approach demonstrates superior accuracy across seven out of the ten parameters compared to alternative methods, as indicated by lower MAE scores. For the remaining three parameters, our performance remains competitive. Specifically, our model achieves the lowest MAE values for parameters $f_x$, $f_y$, $p_x$, $p_y$, $d$, $t_y$ and $\theta_p$ indicating its effectiveness in predicting intrinsic and extrinsic camera parameters. Moreover, our model excels in estimating principle points offset ($p_x$, $p_y$) and pitch angle ($\theta_p$), with notably low MAE values. These findings underscore the robustness and effectiveness of our proposed approach in handling diverse datasets and accurately predicting key parameters for stereo vision tasks.

\subsection{Ablative Study}
We conducted an ablative study (see Table \ref{tab:Ablative Study}), to investigate the impact of variations of geometric constraint loss (UGCL) on our camera calibration model to understand their individual and combined effects on calibration accuracy. The study was structured around three key constraints: UGCL-VP, which incorporates constraints based on vanishing points [\ref{eq:vanishing point}]; UGCL-VP-WC, which extends UGCL-VP by including the projection of the world center [\ref{eq:world center}]; and UGCL-VP-R, which builds upon UGCL-VP-WC by further integrating all the constraints related to the rotation matrix [\ref{sec:Pop Rotation}].\\

\begin{figure}[htb]
    \centering
    \includegraphics[width=0.8\linewidth]{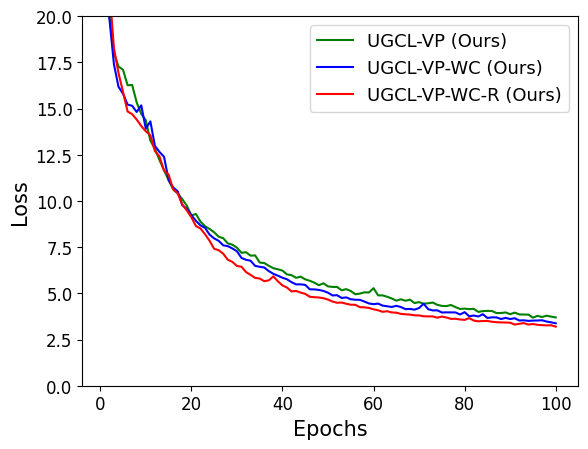}
    \caption{Training loss across 100 epochs with the incorporation of each constraint.}
    \label{fig:ablative}
\end{figure}

The incremental integration of these constraints allowed us to measure the contribution of each geometric consideration to the overall performance of our calibration model. Starting with the basic UGCL-VP setup which implements the vanishing point constraints, we observed a reduction in loss as compared to state-of-the-art, highlighting the importance of vanishing point constraints in our model. Further enhancements to the intrinsic parameter estimation were obtained in the UGCL-VP-WC configuration with the addition of world center projection. The most comprehensive setup, UGCL-VP-R, incorporated rotation matrix constraints offering further performance improvements in the estimation of intrinsic parameters. The effect of adding each constraint can also be seen in Fig.~\ref{fig:ablative}, showing the smoothing of the loss curves when each constraint is added, trained for 100 epochs on the updated CVGL dataset. This progression underscores the cumulative impact of geometric constraints on enhancing camera calibration accuracy, with each additional constraint layer contributing to a more precise and robust calibration model.

\section{Conclusions}
\label{sec:print}

In conclusion, our approach to camera calibration by integrating geometric constraints within a neural network framework marks a significant advancement in precision and interpretability. This results in bridging the gap between traditional camera models and modern machine learning, our method not only achieves superior results across various parameters but also enhances the model's ability to learn and generalize effectively. The incorporation of real-world configurations in an updated CVGL Camera Calibration dataset further reinforces the practical applicability of our work. This study paves the way for a more informed and constrained learning process in camera calibration, contributing to advancements in computer vision and strengthening the robustness of applications reliant on accurate camera parameters.

\clearpage

\bibliographystyle{plain}
\bibliography{reference}

\end{document}